\documentclass{article}
\usepackage{spconf}

\usepackage{graphics}
\usepackage{graphicx}
\usepackage{epsfig}

\usepackage{amssymb}
\usepackage{amsthm}

\usepackage{multirow}

\usepackage{subfigure}
\usepackage{graphicx}
\usepackage{amsmath}
\usepackage{amssymb}
\usepackage{color}
\usepackage{arydshln}
\usepackage{rotating}

\title{Multi-stream 3D FCN with Multi-scale Deep Supervision for multi-modality isointense infant brain MR image segmentation }
\name{Guodong Zeng and Guoyan Zheng}

\address{Institute for Surgical Technology \& Biomechanics, University of Bern, Bern, Switzerland. \\
				Contact: \texttt{guoyan.zheng@istb.unibe.ch}
				}

\begin{document}
%
\maketitle
\begin{abstract}
We present a method to address the challenging problem of segmentation of multi-modality isointense infant brain MR images into white matter (WM), gray matter (GM), and cerebrospinal fluid (CSF). Our method is based on context-guided, multi-stream fully convolutional networks (FCN), which after training, can directly map a whole volumetric data to its volume-wise labels. In order to alleviate the potential gradient vanishing problem during training, we designed multi-scale deep supervision. Furthermore, context information was used to further improve the performance of our method. Validated on the test data of the MICCAI 2017 Grand Challenge on 6-month infant brain MRI segmentation (iSeg-2017), our method achieved an average Dice Overlap Coefficient of 95.4\%, 91.6\% and 89.6\% for CSF, GM and WM, respectively. 
\end{abstract}


\section{Introduction}
\label{sec:intro}
Brain development is complex and spans throughout childhood and adolescence, involving numerous processes such as neural induction, neuronal proliferation and migration, synaptogenesis and myelination, etc. Thus, it is important to develop quantitative tools for analysis of neurodevelopment at all ages. Brain segmentation in MR images is a central piece of such quantitative analysis tools, because it delivers quantitative volume measurement of different brain structures and provides context information for further quantification. An example is the studying of normal and abnormal early brain development where accurate tissue segmentation of infant brain images into white matter (WM), gray matter (GM), and cerebrospinal fluid (CSF) plays an important role.

Despite significant progresses achieved for segmentation of adult brain MR images \cite{Despotovic_CMMM_2015}, the segmentation of infant brain MR images remains a challenge due to ongoing maturation and myelination process in the first year of life \cite{Weisenfeld_NI_2009, Wang_PO_2012}. Moreover, most of the existing infant brain image segmentation methods relied either on the T2 modality for the neonates less than 3 months old \cite{Weisenfeld_NI_2009,Weisenfeld_NI_2009} or on the T1 modality for the infants over 1-year old \cite{Shi_PO_2011}, as in the associated age those modalities demonstrates a relatively good contrast between WM and GM.  Only a few methods \cite{Wang_PO_2012,Wang_NI_2014,Wang_NI_2015,Zhang_NI_2015,Wang_PBMI_2015,Nie_ISBI_2016,Moeskops_CORR_2017} addressed the challenges in segmentation of MR images of isointense-phase infants (around 6-8 months of age). At this stage, T1 and T2 modalities have lowest contrast reflected by the fact that the WM and GM have almost the same intensity level. To address such a challenge, different methods have been proposed before. In \cite{Wang_PO_2012}, Wang et al. proposed a longitudinally guided level set method to segment serial infant brain MT images acquired from 2 weeks up to 1.5 years of age, including the isointense images. To address the difficulty caused by the low contrast, their proposed methods leveraged the complimentary tissue distribution information from 4D longitudinal T1, T2 and diffusion-weighted images. The dependence on the 4D longitudinal data is regarded as the major limitation of this method. To address such a limitation, the same authors later proposed a method to integrate sparse multi-modality representation and anatomical constraint for segmentation of cross-sectional single-time-point isointense infant brain MR images \cite{Wang_NI_2014}. They reported a Dice Overlap Coefficient (DOC) of 0.889 $\pm$ 0.008 for white matter and 0.870 $\pm$ 0.006 for gray matter.   

Recently, machine learning-based methods have gained increasing interest in the field of medical image analysis. Great successes have been validated in different medical image analysis problems. For example, Wang et al. \cite{Wang_NI_2015} proposed a learning-based multi-source integration framework for segmentation of infant brain images. More specifically, they employed the random forest technique to effectively integrate features from multi-source images together for tissue segmentation. More recently, with the advance of deep learning techniques \cite{simonyan2014very,long2015fully,ronneberger2015u}, many researchers have proposed deep learning based methods for automatic infant brain image segmentation \cite{Zhang_NI_2015,Nie_ISBI_2016,Moeskops_CORR_2017}. Both deep convolutional neural networks (CNN)-based methods and fully convolutional networks (FCN)-based method have been introduced before. For example, Zhang et al. \cite{Zhang_NI_2015} proposed a 2D patch-wise CNN to learn a hierarchy of increasingly complex features from T1, T2 and fractional aniostropy (FA) images for the segmentation of multi-modality isointense infant brain image. They showed that their CNN approach outperforms prior methods and classical machine learning algorithms using support vector machine (SVM) and random forest (RF) classifiers. Nie et al. \cite{Nie_ISBI_2016} presented a 2D semantic-wise multi-stream FCN to segment infant brain images using the same datasets that Zhang et al. \cite{Zhang_NI_2015} used. They obtained improved results in comparison to those achieved by Zhang et al. \cite{Zhang_NI_2015}. Their overall DOC were 85.5\% (CSF), 87.3\% (GM) and 88.8\% (WM) vs. 83.5\% (CSF), 85.3\%(GM), and 86.4\% (WM) by \cite{Zhang_NI_2015}. Moeskops and Pluim \cite{Moeskops_CORR_2017} investigated using a dilated triplanar CNN in combination with a non-dilated 3D CNNs for the segmentation of isointense-phase brain MR images.

In this paper, we propose a 3D semantic segmentation method for accurate tissue segmentation of multi-modality isointense infant brain MR images. Our method is based on context-guided, multi-stream 3D FCN, which after training, can directly map a whole volumetric data to its volume-wise labels. Inspired by previous work \cite{3DUNET_2016,Dou_MEDIA_2017}, multi-scale deep supervision is designed to alleviate the potential gradient vanishing problem during training. It is also used together with partial transfer learning to boost the training efficiency when only small set of labeled training data are available. Moreover, context information is used to further improve the performance of our method. 

\section{Data}
\label{sec:Data}
The data used in this study was provided by the 2017 MICCAI grand challenge on 6-month infant brain MRI segmentation \cite{wang2019benchmark}. The training data provided by the challenge organizers consists of T1 and T2 weighted MR images of 10 subjects. The organizers also released test data containing T1 and T2 weighted images of another 13 patients. Thus, in this paper, our method is first trained on the training data and then evaluated on the test data. 

\begin{figure}
\centering
\includegraphics[height=3.0cm]{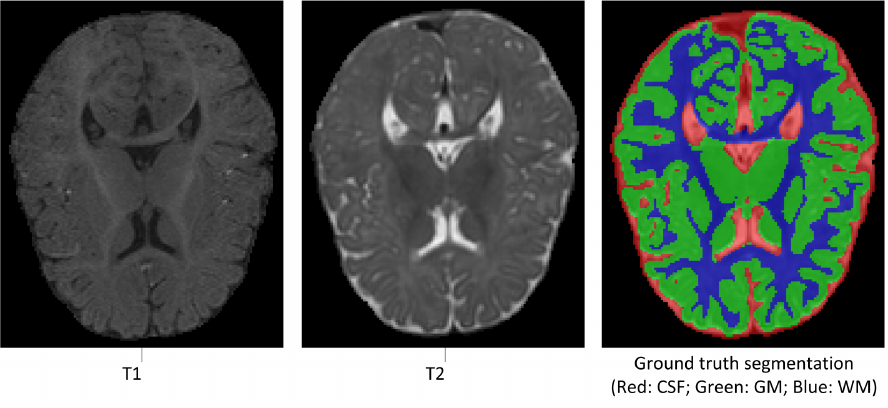}
\caption{ T1 and T2 weighted MR images and the associated ground truth segmentation of a training data}
\label{fig:example}
\end{figure}

All images were preprocessed by the challenge organizers, which included linear alignment of T2 images onto the corresponding T2 images, skull stripping, intensity inhomogeneity correction, and removal of the cerebellum and brain stem. All images were up-sampled into an isotropic grid with a resolution of
$1 \times 1 \times 1 m{m^3}$. Fig. \ref{fig:example} shows the T1 and T2 weighted MR images and the associated ground truth segmentation of a training data

\section{Method}
\label{sec:Method}

\begin{figure}
\centering
\includegraphics[height=5.0cm]{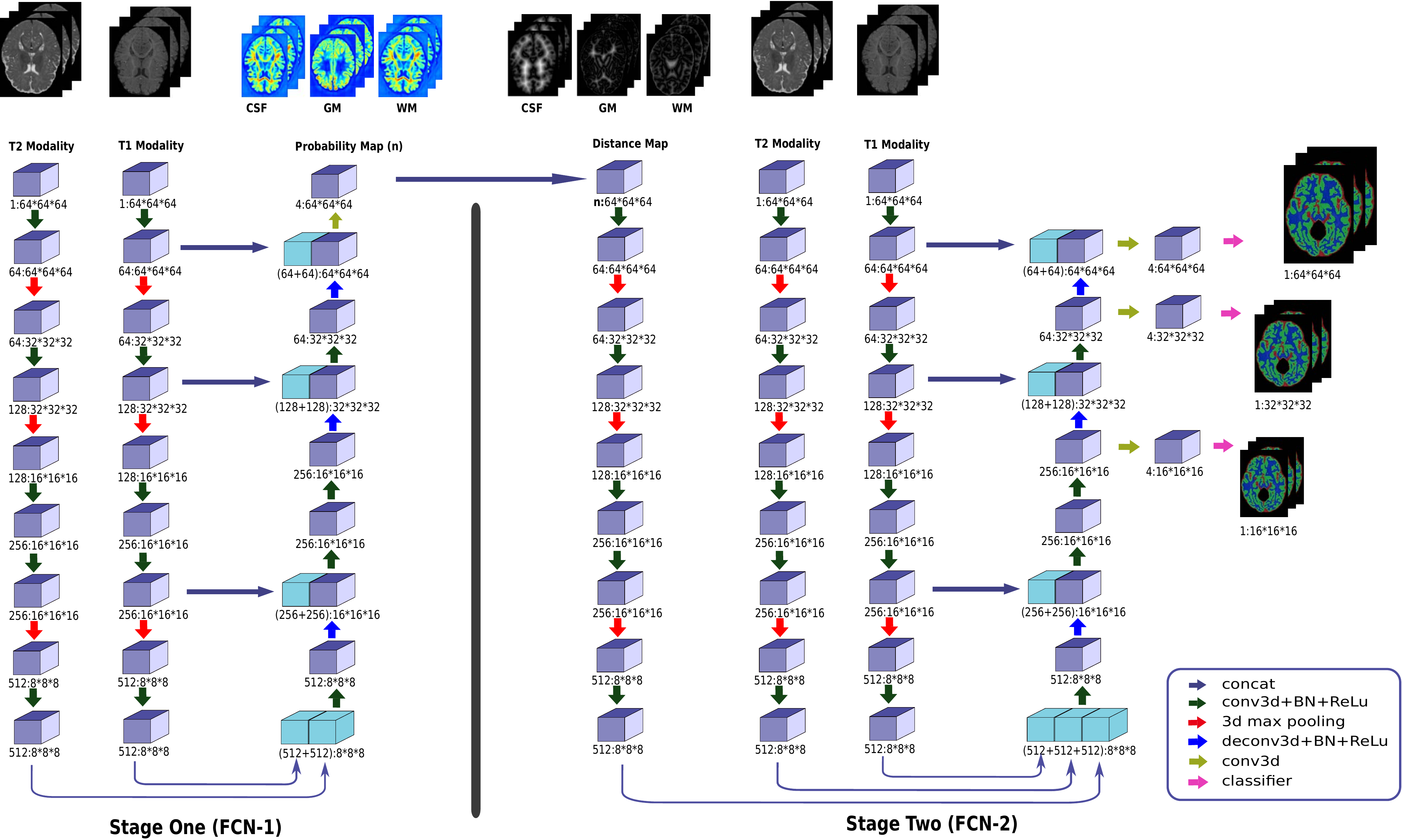}
\caption{A schematic illustration of our proposed network architecture. For each block, the digits below take a format as ``the number of feature stack : the data size"}
\label{fig:network}
\end{figure}

Fig. \ref{fig:network} illustrates our two-stage method for the automatic infant brain segmentation in multi-modality MR images. We first develop FCN-1 which is used at Stage one to learn the probability map of each brain tissues from multi-modality MR images (T1 and T2). An initial segmentation of different brain tissues is then obtained from the probability map, which further allows us to compute a distance map for each brain tissue. The computed distance maps can be used to model the spatial context information. We then develop FCN-2 which is used at Stage two to get the final segmentation by using both the spatial contact information and the multi-modality MR images. In this section, firstly the detailed architecture of our proposed model is elaborated, and then we will introduce the multi-scale deep supervision. Finally, partial transfer learning with is designed to boost the training efficiency, will be described.

\subsection{Multi-stream 3D FCN with Skip Connection}
\label{subsec:Model}

At both stages, multi-stream 3D FCN with long and short skip connections is employed to integrate information from multiple sources, i.e., T1 and T2 weighted images (and context information for FCN-2). More specifically, both FCN-1 and FCN-2 consist of two parts, i.e., the encoder part (contracting path) and the decoder part (expansive path). The encoder part focuses on analysis and feature representation learning from the input data while the decoder part generates segmentation results, relying on the learned features from the encoder part. Unlike previous work \cite{Zhang_NI_2015}, which accommodates multiple sources of information in the form of channels presented to the input layer, we propose to construct an encoder path for each modality and then effectively fuse high-level information from all modalities at the beginning of the decoder path. We feel that the high level information extracted from different modalities at the end of the encoder path are more complementary to each other than the original images from different modalities.

Inspired by 3D U-net \cite{3DUNET_2016}, long and short skip connections, which help recover spatial context lost in the contracting encoder, are used in our network as shown in Fig. \ref{fig:network}. The importance of skip connection in biomedical image segmentation has been demonstrated by previous work \cite{Drozdzal_SkipConnection_2016,ronneberger2015u}.

It has been shown that small convolutional kernels are more beneficial for training and performance. In our deeply supervised network, all convolutional layers use kernel size of $3 \times 3 \times 3$ and strides of 1 and all max pooling layers uses kernel size of $2 \times 2 \times 2$ and strides of 2. In the convolutional and deconvolutional blocks, batch normalization (BN) \cite{ioffe2015batch} and Rectified linear unit (ReLU) are adopted to speed up the training and to enhance the gradient back propagation.

\subsection{Multi-scale Deep Supervision}
Training a deep neural network is challenging. As the matter of gradient vanishing, final loss cannot be efficiently back propagated to shallow layers, which is more difficult for 3D cases when only a small set of annotated data is available. To address this issue, we inject two down-scaled branch classifiers into our network in addition to the classifier of the main network. By doing this, segmentation is performed at multiple output layers. As a result, classifiers in different scales can take advantage of multi-scale context, which has been demonstrated in previous work on segmentation of 3D liver CT and 3D heart MR images \cite{Dou_MEDIA_2017}. Furthermore, with the loss calculated by the prediction from classifiers from different scales, more effective gradient back propagation can be achieved by direct supervision on the hidden layers.

Specifically, let $W$ be the weights of main network and $w$= \{$w^0$, $w^1$, ... $w^{M - 1}$\} be the weights of classifiers at different scales, where $M$ is the number of classifier branches. For the training samples $S$ = {($X,Y$)}, where $X$ represents training sub-volume patches and $Y$ represents the class labels while $Y\in \{0, 1, 2, 3\}$. 

\begin{equation}
  L_{cls}(X,Y;W,w) = \sum_{m=0}^{M-1} \sum_{(x_i, y_i)\in S^m} \alpha_m l^m(x_i,y_i|W,w^{m}) 
\end{equation}
where $S$= \{$S^0$, $S^1$, ... $S^{M - 1}$\}; $S^0$ is a sub-volume patch directly sampled from a training image while $S^m$ contains the examples ($x_i,y_i$) at scale of $m > 0$, which is obtained by downsampling $S^0$ by a factor of $2^m$ along each dimension; $w^{m}$ is the weights of the classifier at scale of $m$; $\alpha_m$ is the weight of $l^m$, which is the loss calculated by a training sample $x_i,y_i$ at scale of $m$.

\begin{equation}
  l^m(x_i,y_i|W,w^{m}) =  {-\log p({y_i} = t({x_i})|} {x_i};W,{w^m})) 
\end{equation}
where $p({y_i} = t({x_i})| {x_i};W,{w^m})$ is the probability of predicted class label $t({x_i})$ corresponding to sample ${x_i} \in S^m$.    

The total loss of our multi-scaled deeply supervised model is then:

\begin{equation}
  L_{total} = L_{cls}(X,Y;W,{w}) + \lambda (\psi (W) \\ + \sum\limits_{m} {\psi (} {w^m}))
\end{equation}

where $\psi()$ is the regularization term (${L_2}$ norm in our experiment) with hyper parameter $\lambda$. 

\subsection{Partial Transfer Learning }
It is difficult to train a deep neural network from scratch because of limited annotated data. Training deep neural network requires large amount of annotated data, which are not always available, although data augmentation can partially address the problem. Furthermore, randomly initialized parameters make it more difficult to search for an optimal solution in high dimensional space. Previous studies \cite{yosinski2014transferable} demonstrated that transferring features from another pre-trained model can boost the generalization, and that the effect of transfer learning was related to the similarity between the task of the pre-trained model and the target task. Furthermore, the same study also demonstrated that weights of shallow layers in deep neural network were generic while those of deep layers were more related to specific tasks. 

To best utilize the advantage of transfer learning, we need to transfer from a model trained on a related task. In this paper, we used a pre-trained model in our previous work \cite{Zeng_MLMI_2017}, which is designed for the task of segmentation of the proximal femur from 3D T1-weighted MR Images. More specifically, the weight of the complete path for T1 modality (including encoder, decoder and all classifiers) are initialized from our previous model \cite{Zeng_MLMI_2017}, while the weights of the encoder path for T2 modality are partially transferred from C3D model \cite{tran2015learning}, which is one of the few 3D models that has been trained on a very large dataset in the field of computer vision.

\section{Experiments and Results}

\subsection{Experimental Setup}
\textbf{Training data augmentation.} Data augmentation was used to enlarge the training samples by rotating each image (90, 180, 270) degrees around the z axis of the image and flipped horizontally (y axis).

\textbf{Training patches preparation.} All sub-volume patches to our neural network are in the size of $64 \times 64 \times 64$. We randomly cropped sub-volume patches from training samples. Each sampled image patch was normalized as zero mean and unit variance before fed into network. 

\begin{table}[t]
\centering
\caption{Table 1. Segmentation performance in terms of DOC, ASD (unit: mm), and MHD (unit: mm) achieved by the present method on the 13 test data.}
\label{Tab:Table1}
\resizebox{8.6cm}{!}{%
\begin{tabular}{llllllllllllllll}
\hline
                                  &  & \#1   & \#2   & \#3   & \#4   & \#5   & \#6   & \#7   & \#8   & \#9   & \#10  & \#11  & \#12  & \#13  & Mean(std)    \\ \hline
\multirow{3}{*}{CSF}                    & DOC     & 0.957 & 0.951 & 0.959 & 0.946 & 0.956 & 0.955 & 0.956 & 0.960 & 0.958 & 0.940 & 0.956 & 0.942 & 0.959 & 0.954(0.007) \\
                                        & ASD     & 0.119 & 0.131 & 0.123 & 0.144 & 0.118 & 0.125 & 0.118 & 0.115 & 0.115 & 0.158 & 0.120 & 0.151 & 0.110 & 0.127(0.015) \\
                                        & MHD     & 7.28  & 9.0   & 11.23 & 9.90  & 9.38  & 11.66 & 8.66  & 9.0   & 8.94  & 11.23 & 7.81  & 10.82 & 10.10 & 9.62(1.35)   \\ \hline
\multicolumn{1}{c}{\multirow{3}{*}{GM}} & DOC     & 0.923 & 0.907 & 0.920 & 0.913 & 0.925 & 0.916 & 0.926 & 0.917 & 0.914 & 0.901 & 0.920 & 0.910 & 0.919 & 0.916(0.007) \\
\multicolumn{1}{c}{}                    & ASD     & 0.30  & 0.361 & 0.337 & 0.351 & 0.318 & 0.333 & 0.280 & 0.320 & 0.335 & 0.428 & 0.321 & 0.411 & 0.333 & 0.341(0.041) \\
\multicolumn{1}{c}{}                    & MHD     & 5.39  & 5.10  & 6.78  & 7.87  & 5.66  & 8.54  & 4.90  & 6.78  & 5.10  & 7.0   & 6.16  & 8.31  & 6.32  & 6.46(1.23)   \\ \hline
\multirow{3}{*}{WM}                     & DOC     & 0.906 & 0.880 & 0.902 & 0.895 & 0.905 & 0.906 & 0.914 & 0.901 & 0.90  & 0.870 & 0.898 & 0.882 & 0.893 & 0.896(0.012) \\
                                        & ASD     & 0.353 & 0.424 & 0.393 & 0.392 & 0.367 & 0.388 & 0.317 & 0.379 & 0.411 & 0.498 & 0.382 & 0.463 & 0.401 & 0.40(0.046)  \\
                                        & MHD     & 7.55  & 6.40  & 8.12  & 7.21  & 5.10  & 7.48  & 4.36  & 7.81  & 7.81  & 6.40  & 6.71  & 7.55  & 5.66  & 6.78(1.16)  
\\ \hline
\end{tabular}
}
\end{table}

\textbf{Training.} We trained our network for $10,000$ iterations after partial transfer learning. All weights were updated by the stochastic gradient descent algorithm (momentum= $0.9$, weight decay=$0.005$). Learning rate was initialized as $1\times10^{-3}$ and halved by every $3,000$ times. In our experiment, we used three branch classifiers. The loss weights of three classifiers $\alpha_0$, $\alpha_1$ and $\alpha_2$ are $1.0$, $0.67$ and $0.33$, respectively. The hyper parameter $\lambda$ was chosen to be 0.005.

\textbf{Testing.} Our trained models can estimate labels of an arbitrary-sized volumetric image. Given images of a test subject, we extracted overlapped sub-volume patches with the size of $64\times64\times64 $, and fed them to the trained network to get prediction probability maps. For the overlapped voxels, the final probability maps would be the average of the probability maps of the overlapped patches, which were then used to derive the final segmentation results. After that, we conducted morphological operations to remove isolated small volumes and internal holes.

\textbf{Evaluation metrics.} For each test subject, automatic segmentation was evaluated against the associated manual segmentation, by using various measurements including DOC, Average Surface Distance (ASD) and Modified Hausdorff Distance (MHD). For details about the evaluation metrics, we refer to the challenge website: http://iseg2017.web.unc.edu/

\subsection{Results}

Table \ref{Tab:Table1} shows the segmentation performance in terms of DOC, ASD and MHD achieved by the present method when evaluated on the 13 test data provided by the 2017 MICCAI grand challenge on 6-month infant brain MRI segmentation. An average DOC of 0.954, 0.916 and 0.896 was achieved for CSF, GM and WM, respectively.

\begin{table}[b]
\centering
\caption{Average results on the test data achieved by the present method when adding context information vs. without adding context information. }
\label{Tab:Table2}
\resizebox{8.5cm}{!}{%
\begin{tabular}{lllll}
\hline
Methods                                                                                        & Tissues & DOC               & ASD               & MHD   \\ \hline           \\
\multirow{3}{*}{\begin{tabular}[c]{@{}l@{}}Adding context \\ information\end{tabular}}         & CSF     & 0.954 $\pm$ 0.007 & 0.127 $\pm$ 0.015 & 9.62 $\pm$ 1.35 \\
                                                                                               & GM      & 0.916 $\pm$ 0.007 & 0.341 $\pm$ 0.041 & 6.46 $\pm$ 1.23 \\
                                                                                               & WM      & 0.896 $\pm$ 0.012 & 0.40 $\pm$ 0.046  & 6.78 $\pm$ 1.16 \\ \hline \\
\multirow{3}{*}{\begin{tabular}[c]{@{}l@{}}Without context \\ information\end{tabular}} & CSF     & 0.950 $\pm$ 0.006 & 0.137 $\pm$ 0.014 & 8.94 $\pm$ 0.98 \\
                                                                                               & GM      & 0.911 $\pm$ 0.008 & 0.366 $\pm$ 0.041 & 6.61 $\pm$ 1.24 \\
                                                                                               & WM      & 0.888 $\pm$ 0.012 & 0.433 $\pm$ 0.050 & 7.12 $\pm$ 1.31 \\ \hline
\end{tabular}
}
\end{table}

\begin{figure}
\centering
\includegraphics[height=5.0cm]{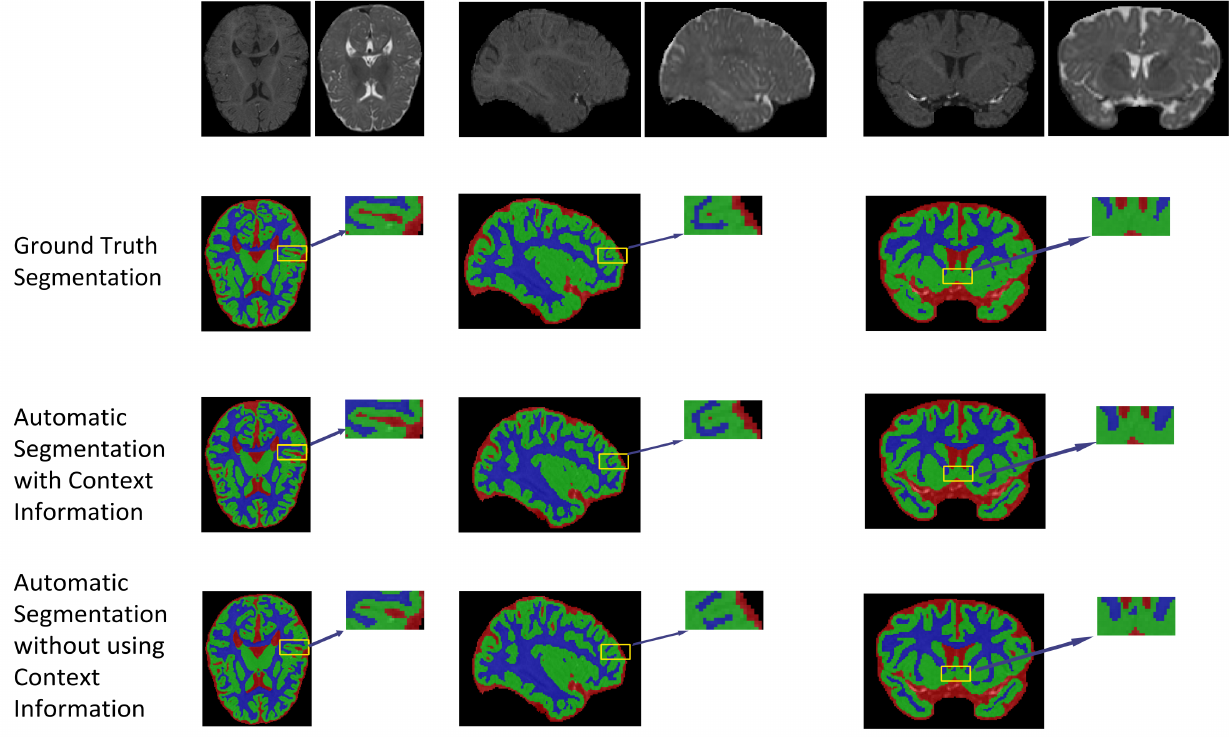}
\caption{A comparison of the ground truth segmentation (the 2nd row), automatic segmentation with context information (the 3rd row) and automatic segmentation without using context information (the 4th row). }
\label{fig:context}
\end{figure}

We also evaluated the effectiveness of adding context information. We compared the results achieved by adding the context information with those achieved without using context information. The results are presented in Table \ref{Tab:Table2}, which clearly demonstrate the effectiveness of adding the context information. Figure \ref {fig:context} shows a qualitative comparison of the automatic segmentation with context information with the automatic segmentation without using context information by taking the ground truth segmentation as the reference.

Implemented with Python using TensorFlow framework and running on a desktop with a 3.6GHz Intel(R) i7 CPU and a GTX 1080 Ti graphics card with 11GB GPU memory, on average our network took about 8 seconds to segment data of one test subject.

\section{Discussions and Conclusions}

In this paper, we proposed a 3D semantic segmentation method for accurate tissue segmentation of multi-modality isointense infant brain MR images into CSF, GM and WM. Our method is based on context-guided, multi-stream 3D FCN with multi-scale deep supervision, which after training, can directly map a whole volumetric data to its volume-wise labels. 

In total 21 teams participated the 2017 MICCAI Grand Challenge on 6-month infant brain MRI segmentation. The performance of all the methods was evaluated and ranked by the challenge organizers. Although our team was placed at the 3rd position out of the 21 teams, the segmentation performance achieved by our method had a very small difference in comparison with those achieved by the teams at the 1st and the 2nd positions. More specifically, the overall DOC achieved by our method and by the team at the 1st position were: 0.954 (CSF), 0.916 (GM) and 0.896 (WM) vs.  0.958 (CSF), 0.919 (GM), and 0.901 (WM).

\bibliographystyle{IEEEbib}
\bibliography{InfantBrainSeg_ISBI2018}

\end{document}